\documentclass{article}
\usepackage{spconf,amsmath,amsfonts,graphicx}
\usepackage{bm}

\usepackage{float,longtable,url,amssymb,lipsum}
\usepackage[ruled,noend]{algorithm2e}
\usepackage[square, sort, numbers]{natbib}
\usepackage{subfigure}
\usepackage{microtype}
\usepackage{booktabs}
\usepackage{multirow}

\title{BRAIN-INSPIRED FEATURE EXAGGERATION IN GENERATIVE REPLAY FOR CONTINUAL LEARNING}
\name{Jack Millichamp, Xi Chen}
\address{Department of Computer Science, University of Bath, Bath, BA2 7PB, UK \\
E-mail: jlm67@bath.ac.uk, xc841@bath.ac.uk}
\begin{document}
%
\maketitle
\begin{abstract}
The catastrophic forgetting of previously learnt classes is one of the main obstacles to the successful development of a reliable and accurate generative continual learning model. When learning new classes, the internal representation of previously learnt ones can often be overwritten, resulting in the model's "memory" of earlier classes being lost over time. Recent developments in neuroscience have uncovered a method through which the brain avoids its own form of memory interference. Applying a targeted exaggeration of the differences between features of similar, yet competing memories, the brain can more easily distinguish and recall them. In this paper, the application of such exaggeration, via the repulsion of replayed samples belonging to competing classes, is explored. Through the development of a 'reconstruction repulsion' loss, this paper presents a new state-of-the-art performance on the classification of early classes in the class-incremental learning dataset CIFAR100.
\end{abstract}
\section{Introduction}
\label{sec:intro}

Artificial neural networks (ANNs) can provide state-of-the-art performance on traditional image classification tasks. Yet, behind this success, there remains the assumption that training inputs are independently and identically distributed (IID). Typically, data is randomly shuffled prior to training, reducing the dependence of consecutive examples. However, this is not always possible; in the real world, samples tend to be temporally correlated, for instance a robot learning new objects classes may first see many images of a dog, then many of a cat. With ANNs, the result of such correlation is a strong interference between learned parameters important to the prediction of current and previously seen classes – causing the model to forget previously learned classes almost entirely. An ideal remedy for such catastrophic forgetting (CF) would involve the storage and interleaved replay of past examples, thus removing any temporal correlation. In practice, however, the storage of an every-growing set of samples is infeasible. Taking inspiration from the brain, a more tractable solution involves the incorporation of a generative model---such as VAE or GAN---into the continual learner, allowing for previously seen classes to be "replayed" alongside the current classes, in much the same way memories are replayed in the mind. The current state-of-the-art generative model, on which this paper builds \cite{van2020}, is able to obtain close to storage based performance on a 10-class MNIST dataset, however, when applied to the more challenging 100-class CIFAR100 dataset, classification precision compared to the stored approach is greatly reduced---with earlier seen classes being affected the most. This paper aims to address the reduction in early class performance by imitating the newfound role of feature exaggeration for the reduction of memory interference in the brain.

\section{FEATURE EXAGGERATION IN THE BRAIN}
\label{sec:brain_exaggeration}

Given the previous success of brain-inspired approaches aimed at mitigating CF \cite{van2020,kirkpatrick2017,zenke2017,shin2017} looking again to the brain---which has developed effective processes to effortlessly tackle such recall---for inspiration seems inevitable. Neuroscience studies have shown that participants trained on competing image-name pairs (such as two similar ape images named Angus and Abner respectively) showed greater repulsion in their neural representations \cite{hulbert2015}. What's more, greater repulsion is correlated with better recall of the image-name pairs, indicating that the brain actively differentiates between highly similar, competing memories to the improve their future recall. A more recent study by Zhao et al \cite{zhao2021} focused on identifying whether the distortion of specific feature dimensions of memories (such as colour) plays a role in memory repulsion. The study found that, given two images of the same object in slightly different colours, participants tended to exaggerate the colour difference between them, such that the pair were more distinguishable. As above, a greater colour exaggeration was associated with a greater recall of linked information. This suggests that the brain adaptively targets and accentuates small differences in important discriminating features of similar memories to reduce interference. Ultimately, this calls to attention to a new approach to deal with interference in ANNs: instead of fighting against interference, the model could be shaped by interference through an adaptive feature exaggeration based approach.

\section{METHODS}
\label{sec:pagestyle}

As described in Section \ref{sec:brain_exaggeration}, the phenomenon of feature-exaggeration in the brain tends to occur in competing memories: where the combination of memories of similar, yet distinct, objects can result in confusion during the recall of either of item. This beneficial memory distortion focuses on exaggerating the differences between the objects, resulting in a repulsion in the brain's internal representations of them which is linked to an improved rate of object recall. In order to mimic this process in the case of continual learning, a measure of competitive classes must first be defined, followed by a method through which the feature-wise differences between these classes can be exaggerated.

\subsection{VAE architecture}
\label{ssec:priorwork}

\begin{figure}[!ht]
        \centering
    \includegraphics[width=8cm]{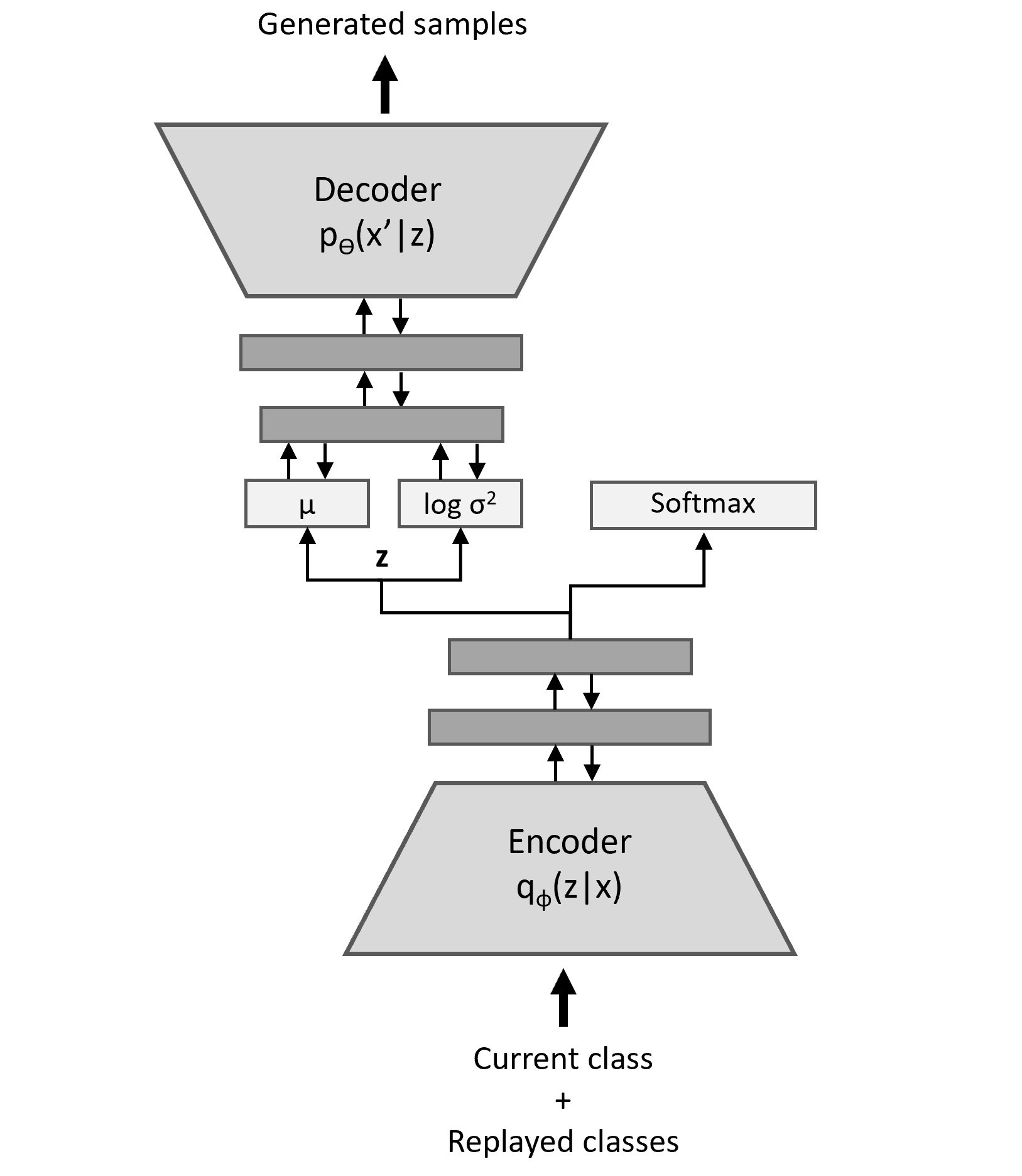}
    \caption{An overview of the BI-R model's architecture \cite{van2020}. Inputs, $\mathbf{x}$, are fed into the encoder and FC layers, converting $\mathbf{x}$ to samples, $\mathbf{z}$, from the conditional distribution $q_{\bm{\phi}}(\mathbf{z}|\mathbf{x})$. Depending on whether classification or sample generation was occurring, the $\mathbf{z}$ samples would then be passed through either the softmax (classification) or the split layer (generation). At the split layer, the samples are re-parameterised into $\bm{\mu}$ and $\log\bm{\sigma^2}$ before passing through two more FC layers and the decoder. $\phi$, $\theta$ are the encoder, decoder weights respectively.}
    \label{fig:model_architecture}
\end{figure}

The model used in this work is built upon the brain-inspired replay (BI-R) model proposed by van de Ven et al \cite{van2020}. This combines a softmax classifier with a VAE to generate samples based on previously seen classes and replay them alongside the current inputs, as depicted in Fig. \ref{fig:model_architecture}. The aim of a VAE's encoder is to distil the high-dimensional input space, $\mathbf{x} \in \mathbb{R}^{n}$, into a lower-dimensional latent space, $\mathbf{z} \in \mathbb{R}^{m}$ (where $m<n$), by approximating the intractable posterior, $p(\mathbf{z}|\mathbf{x})$, as a multivariate Gaussian distribution, $q_{\bm{\phi}}(\mathbf{z}|\mathbf{x})$,  via variational inference. This Gaussian is defined via its mean vector, $\bm{\mu}$, and the logarithm of its variance, $\log \bm{\sigma^2}$, which are returned via a split layer after the encoder and two FC layers, as depicted in Fig. \ref{fig:model_architecture}. The latent variables sampled from this distribution are then passed through two more FC layers and the decoder, which acts as the conditional distribution, $p_{\bm{\theta}}(\mathbf{x^{\prime}}|\mathbf{z})$, returning reconstructed samples, $\mathbf{x^{\prime}}$. The VAE is then optimised over successive batches by minimising the per-batch loss function:

\begin{flalign}
\begin{aligned}
&L(\mathbf{x},\mathbf{x}^{\prime},\mathbf{z};\bm{\phi},\bm{\theta},\bm{\lambda}) = \\
&\sum^{N_{s}}_{i=1}
\left[ \lambda_{r}\ell_{rec}(\bm{x}_i,\bm{x}^{\prime}_i;\bm{\phi},\bm{\theta})\right.
\left.+ \lambda_{d}\ell_{reg}(\bm{x}_i;\bm{\phi},\bm{\lambda}) \right], \label{eq:vae_loss_methods_0}
\end{aligned}
\end{flalign}
where $N_{s}$ is the number of samples per batch; $\bm{\phi}$, $\bm{\theta}$ are the encoder/decoder weights and $\bm{\lambda} = (\bm{\mu},\bm{\sigma^2})$. $\lambda_{r}$, $\lambda_{d}$ are the weights of the reconstruction and regularisation losses (denoted as $\ell_{rec}$ and $\ell_{reg}$ respectively). The reconstruction loss $\ell_{rec}(\cdot)$ of Eq. \ref{eq:vae_loss_methods_0} is a feature-wise binary cross-entropy between generated samples and their original counterparts (Eq. \ref{eq:vae_loss_methods_1}). This loss effectively ensures that the reconstructed samples appear as similar to the original samples as possible. 

\begin{flalign}
\begin{aligned}
&\ell_{rec}(\bm{x}_i,\bm{x}^{\prime}_i;\bm{\phi},\bm{\theta}) =\\
&\mathbb{E}_{\mathbf{\epsilon} \sim \mathbb{N}(\mathbf{0},\mathbf{I})}\left[\sum^{M_{f}}_{f=1}x_{i,f}\log(x^{\prime}_{i,f}) + (1-x_{i,f})\log(1-x^{\prime}_{i,f})\right]. \label{eq:vae_loss_methods_1}
\end{aligned}
\end{flalign}
Where $M_{f}$ is the number of features. The regularisation loss $\ell_{reg}(\cdot)$ aims to maintain as simplistic an approximating distribution to the posterior as possible by minimising the difference between the variational approximation, $q_{\bm{\phi}}(\mathbf{z}|\mathbf{x})$, and a standard normal prior distribution, $p(\mathbf{x})$. To implement this, the loss minimises the Kullback–Leibler (KL) divergence between the approximated posterior and prior distributions.

\begin{flalign}
\begin{aligned}
\ell_{reg}(\bm{x}_i;\bm{\phi},\bm{\lambda}) = D_{KL}[q_{\bm{\phi}}(\mathbf{z}_i|\mathbf{x}_i)||p(\mathbf{z}_i)],\label{eq:vae_loss_methods_2}
\end{aligned}
\end{flalign}
where $D_{KL}[\cdot]$ denotes the KL function that measures the distance between two underlying distributions. 

\subsection{Competing classes}
\label{ssec:competing}

To determine the competing classes in practice, the batch of samples is passed through the model's own classifier: the softmax layer. This returns a discrete probability distribution over all classes the model has so far seen, and can be used to identify competing classes for a given sample according to the model itself. Thus, competing classes are identified as those the model deems most similar and vacillates between during classification. By selecting similar classes this way, the newly proposed feature exaggeration inspired loss can target the more frequently miss-classified classes.

\begin{figure}[htb]
        \centering
    \includegraphics[width=7.5cm]{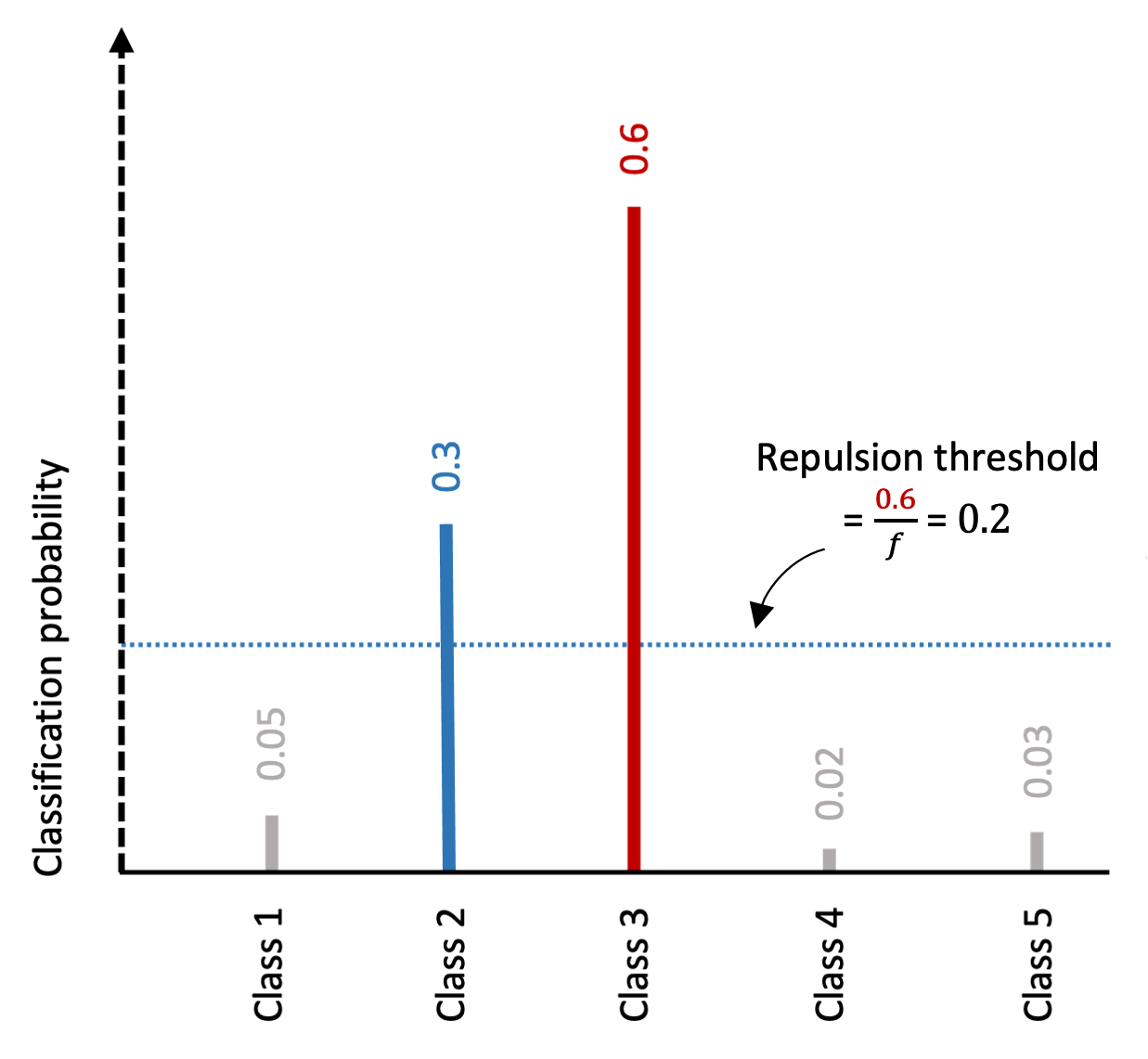}
    \caption{Competing class selection using a repulsion factor. The most likely class (3) is taken as the sample label and its probability is divided by the repulsion factor, $f$, to determine the repulsion threshold. Excluding the label, any class with a probability greater than the threshold is selected as a competing class to the label (2), any below this are ignored. In this example $f$ is 3, making the repulsion threshold 0.2.}
    \label{fig:rep_f}
\end{figure}

To account for varying degrees of competitiveness between classes, an adaptive process, involving the selection of a repulsion factor, $f$, where $f \ge 1$, was used. This factor acts as a dynamic threshold that changes the selection criteria of competing classes depending on the model's confidence in its original label. The highest softmax probability (corresponding to the sample label) is divided by the repulsion factor, providing a threshold probability for identifying competing classes for that sample. Any class with a probability above that of the threshold is deemed competitive. To illustrate this, Fig. \ref{fig:rep_f} depicts the output of the softmax layer for a given sample and process through which competitive classes are identified. If samples exist that don't have any classes other than the label above the competitive threshold, no classes are selected for that sample and the repulsion loss is not calculated over it.

\subsection{Reconstruction repulsion loss}
\label{ssec:RRloss}

To encourage the exaggeration of feature-wise differences between competing classes, a new "reconstruction repulsion" (RR) loss was developed. Taking inspiration from the VAE's reconstruction loss, this acts as a third generative loss, alongside the reconstruction and regularisation terms. It aims to maximise the binary cross-entropy (BCE) between a generated sample, $\mathbf{x}_i^\prime$, and an original input belonging to a competing class $\mathbf{x}_{i_C}$ to that of the sample, as shown in (Eq. \ref{eq:rr_loss}). Through the incorporation of this loss, it is hoped that VAE can be taught to not only learn the characteristic feature vector of a generated class, but to also exaggerate the differences between this class and other competing classes. An important aspect to this, however, is that such repulsion between the features of competing samples can only be done in moderation. Too much and the generated samples may be unrecognisable from their intended classes. Therefore, when combined with the original losses, the RR loss was weighted with a factor, $\lambda_{RR}$, and the ratio between this factor and that of the original reconstruction loss was explored. Furthermore, once the competing classes to a given sample have been determined, a mean feature vector, $\bm{\hat{x}}$, for each competing class is calculated via the mean of all original samples in that class. Having the same shape and dimensionality as the inputs, this mean feature vector can be viewed as a generalised sample that can capture more of the characteristic feature patters of a given class.

\begin{flalign}
\begin{aligned}
&\ell_{RR}(\bm{x}^{\prime}_i, \bm{\hat{x}}_{i_C};\bm{\phi},\bm{\theta}) =  \\ 
& \mathbb{E}^{-1}_{\mathbf{\epsilon} \sim \mathbb{N}(\mathbf{0},\mathbf{I})}\left[\sum^{M_{f}}_{f=1}\hat{x}_{i_{C},f}\log(x^{\prime}_{i,f}) + (1-\hat{x}_{i_{C},f})\log(1-x^{\prime}_{i,f})\right]\label{eq:rr_loss} 
\end{aligned}
\end{flalign}

In addition, a similar reconstruction attraction (RA) loss was also used to exaggerate the features of replayed samples belonging to the same class towards each other, in an effort to attract their internal representations (Eq. \ref{eq:ra_loss}). The reasoning behind this relates again to Zhao et al's \cite{zhao2021} observations of memory repulsion in the brain---namely that memories of related objects tend to be exaggerated towards each other.

\begin{flalign}
\begin{aligned}
&\ell_{RA}(\bm{x}^{\prime}_i, \bm{\hat{x}};\bm{\phi},\bm{\theta}) = \\ &\mathbb{E}_{\mathbf{\epsilon} \sim \mathbb{N}(\mathbf{0},\mathbf{I})}\left[\sum^{M_{f}}_{f=1}\hat{x}_{f}\log(x^{\prime}_{i,f}) + (1-\hat{x}_{f})\log(1-x^{\prime}_{i,f})\right]  \label{eq:ra_loss}
\end{aligned}
\end{flalign}

Due to the fact that the exaggeration observed in the brain only occurs between memories themselves and not to the objects responsible for these memories, the RR and RA losses were only applied to batches of replayed samples, rather than new class batches.

\section{RESULTS}
\label{sec:typestyle}

\begin{table}
\centering
\begin{tabular}{p{1cm} l p{1.6cm} r p{1.6cm} r p{1.6cm} r}
 \toprule
 \multirow{2}{*}{$f$} & \multicolumn{3}{c}{Improvement over baseline (\%)} \\
 & 20\% avg. & 50\% avg. & Overall avg. \\
 \midrule
 5 & 3.80\slash12.5 & 4.06\slash12.78 & 1.35\slash3.87 \\
 10 & \textbf{5.80\slash19.08} & 3.82\slash12.03 & 1.19\slash3.41 \\
 20 & 4.80\slash15.79 & \textbf{4.86\slash15.30} & \textbf{1.48\slash4.25} \\
 50 & 4.90\slash16.12 & 3.56\slash11.21 & 1.26\slash3.61 \\
 100 & 4.95\slash16.28 & 4.18\slash13.16 & 1.24\slash3.56 \\
 \bottomrule
\end{tabular}
\caption{\label{tab:CL_soft} Table showing the absolute and relative change in precision above that of the baseline for various repulsion factors, in the format: absolute change/relative change. Comparisons are shown across the average (denoted as avg.) of all test classes (Overall), the first 50\% of classes seen and the first 20\% seen. Experiments were run using 5000 iterations with the pre-determined optimal hyperparameters.}
\end{table}

\begin{figure}[!ht]
    \centering
    \includegraphics[width=8.5cm]{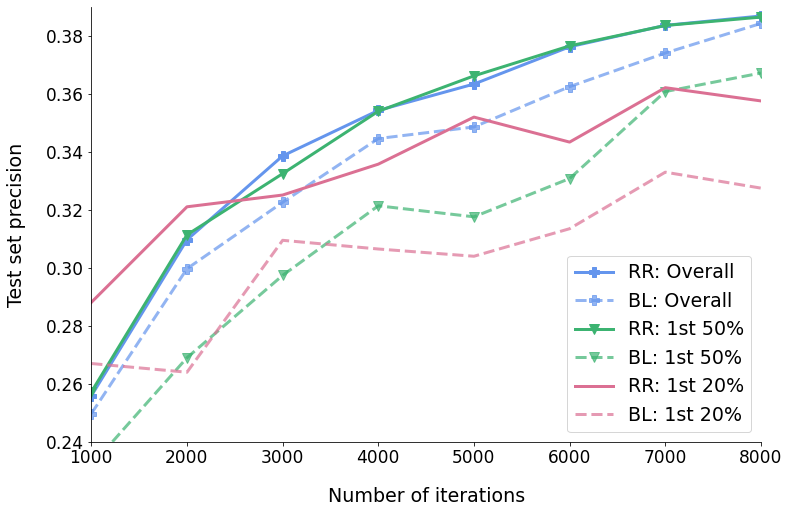}
    \caption{A comparison of the final classification test set precision and convergence rates of the RR+RA adaptation (RR) and baseline (BL). Presented are the overall precision (average across all classes; a), precision across the first 50\% of classes encountered (b) and over the first 20\% of classes encountered (c). This was performed over current and replay batch sizes of 512, with $\lambda_{RR}$ and $\lambda_{RA}$ weightings of $10^{-6}$ (the optimal values).}
    \label{fig:RR_covergence}
\end{figure}

The final performance of the RR+RA model with optimised hyperparameters was compared to that of the state-of-the-art BI-R "baseline" model, using the optimised hyperparameters as stated by van de Ven et al \cite{van2020}. Given that CF directly reduces the final classification precision of previously seen classes, this was main metric employed as an indicator of continual learning performance. All results stated relate to the average test precision across 5 runs of the experiment with optimal hyperparameters. Due to the nature of CF, it will tend to have a greater detrimental effect on the learned representations of earlier-seen classes, whose internal representations are more likely to be 'over-written'. Thus, the average precision on the first 50\% of classes seen will tend to be lower than the average over the last 50\% and any successful method will likely lead to an improvement, not necessarily in the overall average precision, but in the average precision of the first 50\% and most likely the first 20\% of classes.

The final precision and convergence rate of the RR+RA adaptation was compared to the baseline over a range of gradient descent iterations, as shown in Fig. \ref{fig:RR_covergence}. The average precision across all classes (Fig. \ref{fig:RR_covergence}a) showed a slight, although not significant, improvement across the tested iterations. Comparing the average precision of the first 50\% of classes (Fig. \ref{fig:RR_covergence}b), however, the inclusion of the RR+RA loss provided a significant increase over the baseline for all tested iterations. For instance, the RR+RA model achieved an absolute increase in percentage precision of 3.20\% and a relative increase of 15.30\% when compared to baseline for 5000 iterations. The average performance over the first 20\% of classes seen was also compared to the baseline, showing similar significant improvements to classification precision across all iterations. A slightly larger absolute improvement of 4.30\% and relative increase of 15.79\% over the baseline was obtained for 5000 iterations. The best values for $f$ proved to be 10 and 20 (Table \ref{tab:CL_soft}). Performance using the repulsion factor, $f$, as described in \ref{ssec:competing} was also compared to that without it without it, with the inclusion of $f$ providing a mean relative precision improvement of 3.45\% for the first 50 percent of classes.

\section{DISCUSSIONS AND CONCLUSIONS}
\label{sec:discussion}

For all average metrics used, the RR+RA loss provided a significant improvement over the baseline classification precision. This improvement was amplified across the classification of the earlier-seen classes, with a mean relative precision improvement of 15.30\% for the first 50 percent of classes and 15.79\% for the first 20 percent of classes (absolute changes of 4.86\% / 4.8\%). These strong early class results act to confirm the hypothesis that a targeted feature-based exaggeration of samples allows the model to more easily learn the characteristic differences and similarities between samples of competing and non-competing classes, resulting in a reduction of CF amongst earlier-learnt classes. To further confirm the internal repulsion mechanism through which this adaptation improves precision, future work could focus on analysing the effect of the loss on the internal representations of competing classes.

The RR+RA adaptation is certainly not perfect---relying on the assumption of correct class labels being determined during the generation of replayed batches. This assumption does not always hold true and could be a source of interference in the optimisation of the RR loss. Further research could aim to incorporate the model's confidence in its predictions in the RR loss. This alteration has the potential to lead to a further boost in classification performance.

Taking a step back, this paper is centred around the task of continual image classification, but modifications such as the RR+RA loss could also be applied to other continual learning contexts with different data-types. For instance, given that the loss only depends on the BCE between the feature-vectors of competing samples, as long as the data can be translated into feature-vector form this could be calculated regardless of the vector's shape or dimensions.

\vfill\pagebreak

\bibliographystyle{IEEEbib}
\bibliography{strings}

\end{document}